# Explainable Unsupervised Multi-Anomaly Detection and Temporal Localization in Nuclear Times Series Data with a Dual Attention-Based Autoencoder

Konstantinos Vasili,[a] Zachery T. Dahm,[a] and Stylianos Chatzidakis[a]*

[a]*Purdue University, School of Nuclear Engineering, West Lafayette, IN*

*E-mail: schatzid@purdue.edu

# Explainable Unsupervised Multi-Anomaly Detection and Temporal Localization in Nuclear Times Series Data with a Dual Attention-Based Autoencoder

**Abstract**: The nuclear industry is advancing toward more new reactor designs, with next-generation reactors expected to be smaller in scale and power output. These systems have the potential to produce large volumes of information in the form of multivariate time-series data, which could be used for enhanced real-time monitoring and control. In this context, the development of remote autonomous or semi-autonomous control systems for reactor operation has gained significant interest. A critical first step toward such systems is an accurate diagnostics module capable of detecting and localizing anomalies within the reactor system. Recent studies have proposed various ML and DL approaches for anomaly detection in the nuclear domain. Despite promising results, key challenges remain, including limited to no explainability, lack of access to real-world data, and scarcity of abnormal events, which impedes benchmarking and characterization. Most existing studies treat these methods as black boxes, while recent work highlights the need for greater interpretability of ML/DL outputs in safety-critical domains. Here, we propose an unsupervised methodology based on an LSTM autoencoder with a dual attention mechanism for characterization of abnormal events in a real-world reactor radiation area monitoring system. The framework includes not only detection but also localization of the event and was evaluated using real-world datasets of increasing complexity from the PUR-1 research reactor. The attention mechanisms operate in both the feature and temporal dimensions, where the feature attention assigns weights to radiation sensors exhibiting abnormal patterns, while time attention highlights the specific timesteps where irregularities occur, thus enabling localization. By combining the results, the framework can identify both the affected sensors and the duration of each anomaly within a single unified network.

## I. INTRODUCTION

The operation of a nuclear reactor historically relied on human operators to perform all basic operations and system diagnostics. Once implemented, new advanced designs are envisioned to rely on digital instrumentation and potentially remote operation via autonomous or semi-autonomous control systems [1][2], aiming to minimize human intervention and reducing costs. An autonomous control system is described as an intelligent system with ability to self-perform and execute control functions under uncertain conditions [3]. Despite recent advancements [2][4][5][6], a fully operational autonomous control nuclear system is still far from its intended performance. The degree of autonomy in such a system, depends upon the extent to which it can reliably perform tasks such as fault diagnosis, self-validation, forecasting, and decision-making with minimum human intervention [1] [3].

The uncertainty and risk associated with the operation of nuclear systems, makes the design of autonomous control a challenging task. Studies in recent years have examined various techniques for different modules of an autonomous or semi-autonomous control system. Initial studies relied on mathematical or statistical approaches to simulate autonomous control modules [7]. Today, the growing interest in Artificial Intelligence (AI) and Machine Learning (ML) methods highlights their potential for various nuclear applications and aligns with the industry's drive for advanced data analytics techniques integrated in new reactor designs. This growing interest is reflected in the number of studies that leverage advancements in AI and ML to monitor, predict critical infrastructure failures, perform accident diagnosis, and manage reactor operations [5][6] [8][9][10] [11].

In [5] a fault detection model that employs Deep Learning (DL) to enhance the detection of faults in nuclear power plants is proposed. The DL models were tested on a simulation of the IP-200 nuclear power plant using RELA5/MOD 4 [12] evaluating its effectiveness in diagnosing faults. In [6] the authors propose an online Deep Neural Network model for real-time fault monitoring using again RELAP5-generated data. The proposed framework could monitor and detect faults in operational settings, which could lead to improved safety and reduced downtime in nuclear facilities. In [11] a Long Short-Term Memory (LSTM) Autoencoder (AE) algorithm is proposed for identifying accidents based on simulated historical data. The study shows that the algorithm can detect accidents and differentiate between trained and untrained events, providing operators with information during emergencies. In [9] the authors attempt to develop a Deep Rectifier Neural Network (DRNN) model to classify various operational scenarios and accident conditions within a nuclear power plant using data from a PWR simulator. The study showed that the DRNN outperforms traditional methods in terms of accuracy and speed, making it a potentially useful tool for enhancing safety and operational reliability in nuclear power plants. A more recent example includes [13], which covers various ML algorithms and their potential to improve monitoring, fault detection, and predictive maintenance in nuclear settings. The study concludes that ML algorithms are well-suited for damage detection in nuclear facilities due to their ability to learn complex patterns in large datasets generated by sensors that monitor structures and systems. However, it also points out significant challenges, including limited real data availability, lack of validation in real-world conditions, and increased computational resources required for the development and deployment of these AI-based frameworks.

Despite the potential of the proposed methods, the complexity of reactor systems, the need for real-time decision-making, and the handling of vast amounts of sensor data present significant challenges that researchers are actively addressing. In [14] the authors discuss the complexity of implementing prognostics and health management in nuclear power plants, highlighting that the integration of various advanced technologies, such as ML and DL for big data analytics, into a cohesive system is inherently challenging. These challenges include data acquisition, real-time processing, model accuracy, and the integration of multiple subsystems into a reliable and robust autonomous control framework. In addition, research conducted in relevant studies [15] [16] [17] has shown that the lack of explainability in advanced data analytics techniques, including AI/ML, restricts their adaptation in critical processes. AI/ML predictions are often regarded as black boxes, as their non-linear structure makes it difficult to determine which inputs cause them to generate particular outputs.

In this work, we propose a framework to address challenges on anomaly detection, including the lack of real-world data, the scarcity of abnormal events in nuclear processes, and the explainability of ML methods, as part of a diagnostics module for an autonomous control system of a real nuclear reactor. In prior results [18], we developed a dedicated unsupervised network to monitor each sensor and generate system diagnostic messages. Building on this work, we now develop an unsupervised learning methodology with a dual attention mechanism to not only detect anomalous sensor readings, but also identify the duration of the anomaly and localize the erroneous sensors within an array of sensors, all within a single deep learning model. We validate our approach with real data from PUR-1 research reactor at Purdue University. We were able to successfully determine the duration of the irregularities, and source localization in four erroneous radiation sensors and characterize scenarios of increased complexity: a) drifted sensor readings, b) anomalous spikes in individual sensors, and c) concurrent and overlapping anomalous readings.

## II. BACKGROUND
### II.A. Monitoring and Diagnostics Module

An accurate diagnostics module is a significant part of an advanced reactor autonomous control architecture for maximizing safety and reducing operational cost [19] [20] [21]. The main functionality involves monitoring different reactor components by receiving inputs from sensors that are mounted in the reactor facility (temperature, neutron fluxes, radiation levels, etc.), identifying the current mode of the reactor system, and providing diagnostics credibility through fail mechanisms [20].

The integration of multiple sensors within and around a reactor provides a valuable source of real-time, multivariate, and time-dependent data for both operational states and system performance. Whenever a fault occurs in a reactor component, the corresponding sensor readings show unique time-dependent variations that the module is expected to be able to analyze and report. Unlike existing automation, where safety alerts are triggered by setpoint activations, an advanced diagnostics module utilizes advanced data analytics that run consecutively within the control loop, either under or in parallel with other reactor operation processes, to determine whether a failure metric is present.

The concept of applying AI/ML methods to simulate the operation of a diagnostic module has gained attention among researchers [18] [19]. The main difference in our study is that a single network is used to monitor simultaneously an array of sensors and identify the root cause by localizing the sensors that trigger the abnormality without the need to define irregular classes manually. The high dimensionality of the signal space in modern reactor systems, where up to 2000 features per sampling frequency can be monitored, makes it impractical to assign a dedicated model to monitor each signal or to rely solely on supervised learning to capture all potential irregularities within the system. In this context, the control loop architecture becomes more efficient by grouping sensors of similar behaviour (e.g. radiation detectors, neutron flux detectors, etc.) and dedicating each group to a monitoring network. This way, the number of AI/ML models required is reduced, and the computational complexity of the overall architecture becomes less resource-intensive. To assess the feasibility of the proposed architecture, we utilize the four radiation detectors present in the PUR-1 operation room, as part of a diagnostics module with unsupervised learning capabilities. The results show success in identifying the erroneous readings, along with the duration and the root cause within a single model.

## II.B. Anomaly Detection

The main functionalities of a diagnostics module can be regarded as an anomaly detection task, and it is normal to employ proven solutions from the anomaly detection field. The term anomaly in such an analysis refers to irregular or unusual events that occur with a very low probability [22]. The main distinction in solving anomaly detection problems is between model-driven and data-driven approaches. Model-driven approaches require *a priori* system knowledge, typically expressed through mathematical modeling with specific parameterizations of system dynamics, which can be difficult to obtain for complex non-linear systems. Data-driven approaches rely on the availability of a vast amount of data to capture the system dynamics. Data-driven approaches are further subdivided in supervised and unsupervised methods.

Supervised methods, where the classes are explicitly defined in the training data, have been extensively studied. Classical approaches for time series anomaly detection include methods such as distance-based in early [23] and recent years [24], in which the explicit distance between two temporal sequences is used to quantify the similarity between the two different classes. Clustering-based approaches [25], which classify the data into distinct classes, predictive-based [26], and probabilistic-based [27], [28] techniques. While these methods are computationally simple, they struggle to capture complex structures within high-dimensional data and often require input from domain experts, which restricts their generalizability. AI/ML algorithms have been proposed to resolve these limitations. Algorithms that have been successfully applied to time series data include linear regression, logistic regression, support vector machine, k-nearest neighbour, decision trees, random forest, isolation forest, ARIMA, and Neural Networks [29] [30] [31] [32] [33] [34] [35]. More recently, DL techniques have been shown to offer advantages over traditional AI/ML methods. They provide end-to-end solutions and are well-suited to overcoming these limitations due to their inherent ability to learn complex relationships from both high and low-dimensional data without requiring in-depth domain knowledge [36]. Comprehensive reviews of the available methods can be found in these studies [37] [38].

### II.C. Unsupervised methods for Anomaly Detection

Despite the success of the supervised methods, a key restriction is that labelled data are required for every training class and in sufficient amounts. In well-optimized systems, like in the case of a nuclear reactor, non-anomalous data are abundant, whereas abnormal data are rare, making supervised techniques impractical. To overcome labelling, scarcity, or class imbalance, unsupervised methods have been developed. Similar to the supervised methods, classical unsupervised ML approaches utilize probabilistic, distance-based, clustering or dimensionality reduction methods [39]. Algorithms associated with these techniques include k-means, DBSCAN, Gaussian Mixture Models, and PCA. However, they have limited capability to capture complex structures as data volume or feature space increases. To address these challenges, unsupervised DL methods, such as Autoencoders (AEs) [40] and Generative Adversarial Networks (GANs) [41], along with CNN and LSTM-based variants, have emerged.

GANs were first introduced in the work of Goodfellow et al., 2014 [42]. They consist of two neural networks, called the generator (G), which produces samples that mimic the distribution of the training data, and the discriminator (D), which determines between real and generated samples. During training, a minmax objective function is optimized, where D is trained to maximize its ability to discriminate real from fake samples, while G is trained to generate samples that D cannot distinguish from real data, until an equilibrium is reached. AEs were initially introduced by Rumelhart et al., 1986 [43] for efficient image storage and transmission through dimensionality reduction of the input data. A typical architecture consists of three components: the encoder, the latent space, and the decoder. The encoder uses a series of layers to compress the input data into a lower-dimensional representation, called the latent space, which captures the most essential features in the data. The decoder reconstructs the data to its original dimensions. The network learns to optimize its parameters during training by minimizing the reconstruction error between the input and the reconstructed data. Unsupervised methods using AE have shown strong performance for anomaly detection in multivariate time series (MTS) data. The main concept is that the network can be used to evaluate unseen data, where the high reconstruction error indicates a deviation from expected patterns and can be regarded as a potential irregularity of the system.

### II.D. Autoencoder for anomaly detection in Nuclear

AEs have been used in a number of studies in the nuclear domain. This study [44], explores the use of AEs to detect anomalies in simulated nuclear data. A pressurized water reactor model was used to generate data points consisting of a number of reactor features. A portion of the datasets was modified by introducing different levels of Gaussian noise, and an AE architecture was trained to distinguish normal from noisy data based on the reconstruction error. However, the study does not consider the time dimension and therefore, is not applicable for real-time monitoring. Following the same approach, this study [45] introduces an efficient AE methodology for predicting potential anomalies in the cooling system of a simulated PWR, based on the reconstruction error between unseen and trained data. Similar studies include [46] [47], where a residual error analysis is used to indicate potential anomalies in their specific cases. In [48], the authors investigate an LSTM-AE network for MTS and extend the concept of residual analysis by employing the Mahalanobis distance to quantify the overall model residual, which serves as a metric to determine the anomaly.

As described in the above studies, the primary focus of applying AEs in the nuclear field relies on reconstruction error analysis to identify general anomalies within the system. However, AEs, especially those consisting of fully connected layers, are known to suffer from data leakage. AEs learn correlations across all input dimensions, meaning that a large error in one sensor can propagate error influence on other sensors. As a result, it becomes challenging to identify the root cause or determine the duration of an anomaly in a multisensory environment without input from a domain expert [49]. In addition, due to the nature of nuclear data and the inherent uncertainty in the measurements, the reconstruction error of abnormal events may fall within the statistical variability of normal events, potentially leading to high error rates (false alarms). To address this limitation, attention mechanisms [50] have recently been shown to be an effective enhancement to DL architectures [51], enabling the adaptive weighting of feature contributions. Despite their success in supervised anomaly detection frameworks [52][53][54] their application for unsupervised anomaly detection, particularly in the nuclear domain, is limited. Here, we investigate an attention mechanism as part of an AE architecture as an alternative to traditional reconstruction error analysis for detecting and localizing anomalies. The AE architecture consists of layers with time-learning capabilities, such as LSTM networks [55], which are well-suited for time series data due to their ability to capture long and short-term temporal dependencies. Unlike traditional neural networks or recurrent neural networks, LSTMs are designed to retain information over large sequences, making them ideal for modelling data with time-dependent patterns, such as reactor sensor measurements.

*II.E. Attention Mechanism*

The attention mechanism was first introduced by Bahdanau et al., 2015 [56] in the context of natural language processing, and specifically for machine translation. Before attention was proposed, machine translation typically relied on encoder-decoder architectures. These models read sequences in one language (e.g., English), compress them into a fixed-length vector, and generate the translated output sequence (e.g., French). However, when dealing with long sequences, the decoder often struggled to retain relevant information from earlier parts of the sequence, resulting in poor performance.– The attention mechanism addressed this limitation by allowing the decoder to selectively focus on different parts of the input sequence. Instead of treating all input elements equally, the model learns to assign greater weights to the most relevant parts. Given an input sequence:

$$X = [x_1, x_2, \ldots, x_T]$$

Where T is the sequence length and $x_i \in R^d$ (d is the dimensions of the vectors). A recurrent neural network, such as an LSTM or GRU, produces a sequence of hidden states per step:

$$H = [h_1, h_2, \ldots, h_T]$$

Where each $h_t$ is a latent representation that summarizes the information up to time *t*. In additive (Bahdanau) attention, a context vector for each target position i in the sequence X is computed as the weighted sum of these $h_i$ states:

$$c_i = \sum_{j=1}^{T} a_{ij} h_j$$

Where $a_{ij}$ are the attention weights obtained via a softmax over the alignment scores:

$$a_{ij} = \frac{exp(e_{ij})}{\sum_{k=1}^{T} exp(e_{ik})}$$

The alignment scores are defined as:

$$e_{ij} = a\,(s_{i-1}, h_j)$$

Where $s_{i-1}$ is the decoder state at the previous step, and $a\,(\cdot)$ is a learned function that measures how well the inputs around position *j* align with the output at position *i*. In practice, by applying the attention mechanism to the input sequence, the model constructs a context-specific summary of the input for each output step. This enables the decoder to focus on the most relevant input at each step, resulting in improved translation accuracy.

**III. METHODOLOGY**

The proposed methodology consists of an AE architecture with time sequence learning capabilities, which serves as a reconstructive model for selected PUR-1 signals. The data are retrieved from an online monitoring platform and preprocessed to ensure compatibility with the architecture. The dual attention mechanism provides local and global interpretability, enabling the identification of erroneous sensors, the time of occurrence, and duration. An overview of the methodology is shown in Figure 1.

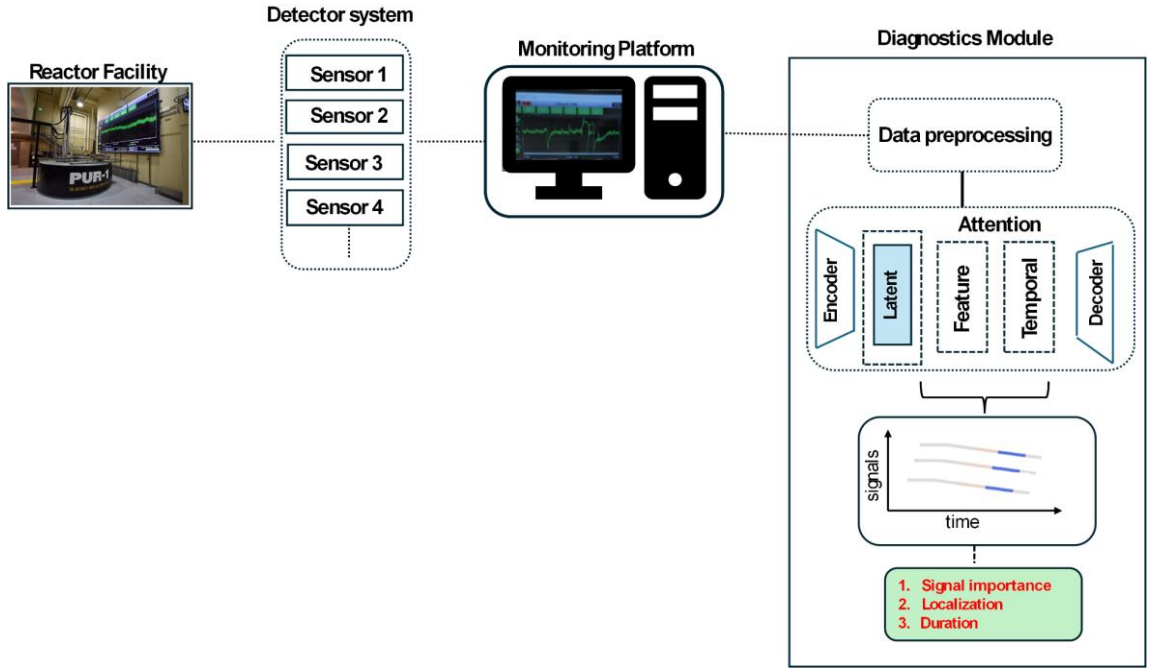

Fig. 1: Methodology overflow

*III.A Data Generation and Collection*

In contrast to simulated data, real operational data are subject to various artifacts, such as fluctuations, outliers, null values, and noise. These inconsistencies arise due to factors, like limitations or imperfections in the data collection instruments, along with the inherent complexity of the monitoring system. In addition, human operation and decision-making behaviors contribute to data variability, as each operator may handle the reactor differently under the same circumstances, which is reflected in the recorded data. These challenges underscore the importance of utilizing real-world data as inputs to data-driven algorithms, as near-perfect synthetic data may not accurately capture the complexities and uncertainties inherent in real reactor conditions.

The data used for this study were extracted directly from the PUR-1 reactor (Fig. 2). PUR-1 is a pool-type research reactor located at Purdue University, unique in its kind as the only research reactor in the United States with a fully digitized Instrumentation and Control system. Several sensors are mounted inside and around the reactor pool, continuously monitoring the reactor state and the surrounding environment during operational and non-operational activity. A digital twin of the reactor has been implemented for research purposes (Fig. 2), featuring a remote monitoring software that enables the collection of real-time data on more than 2000 features per sampling frequency. The data collection is performed by querying the main server that stores the data from the reactor sensors. The queries are formulated with user input commands to filter the data based on a variety of selections (dates, signals, and combinations). The data are retrieved as MTS organized in tabular format, where each row contains a sampling frequency (e.g., 1 second of measurements) and the columns contain the selected signals.

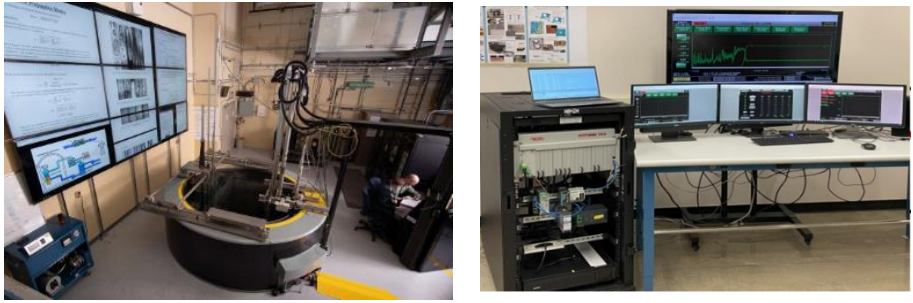

Fig. 2. PUR-1 facility (left). Digital twin and remote monitoring software of PUR-1 (right).

PUR-1 contains four radiation detectors called Radiation Area Monitors (RAMs). Their primary purpose is to monitor sensitive areas of the reactor control room and alert for high radiation doses and contamination. Each RAM detector is a gas-filled Geiger-Muller (GM) detector, known for its high sensitivity to all kinds of radiation. The detectors are distributed across four different locations; the first monitors the reactor pool level (*ram pool*), the second is in the water infiltration system (*ram wtr*), and the third is on the reactor control system (*ram con*). The last is a Continuous Air Monitor (*cam*), which is part GM detector and part proportional counter, monitoring radiation levels in the control room's surrounding environment. The distribution of the detectors across the reactor operation room is presented in Figure 3.

The sensor readings are measured in counts per second, where higher values indicate increased radiation levels. A high radiation reading may indicate either a potential anomaly in the detector or a contamination incident due to radiation leakage. For safety reasons, a low activity source is mounted inside the detectors, which continuously provides a minimum threshold of radiation readings. The problem falls into determining the threshold value above which a sensor reading is considered an abnormal event. The trivial solution of defining a single threshold per sensor is not directly applicable in the current multivariate system, where correlations exist between each sensor and other signals monitoring the reactor. This suggests that an anomaly is part of a higher-dimensional signal space, where each signal contributes to the radiation readings. In this context, the anomaly is expressed as high or small sensor readings in correlation with the readings of other relevant signals at the same time. For example, a large sensor value cannot occur when the reactor power is close to zero. To capture the correlations between sensors and other reactor features, the dataset used for training includes the four radiation detectors along with two signals measuring reactor power (*neutron counts*) and neutron flux (*neutron flux*). Figure 4 presents an example of such a correlation between the *ram pool* and the *neutron counts* detectors during multiple reactor operations concatenated over a period of 30,000 seconds. To facilitate comparison, the neutron count readings are shown on a logarithmic scale. A clear positive correlation can be observed between the reactor power and the radiation levels at the pool. This correlation provides essential information for the model. During training, the model is expected to learn this pattern, and at inference, any deviations significantly higher or lower than the expected behavior may be flagged as potential anomalies.

The other radiation detectors do not exhibit the same strong correlation with the reactor power, likely due to their locations being away from the reactor pool. Nevertheless, we include the entire multivariate dataset to train the network, obtained from a variety of reactor operations at different reactor power levels. This will help capture even subtle correlations that may not be immediately apparent in the data. Moreover, it supports our goal of developing a multisensory diagnostics module using a single network. The signals used for the training, along with a brief description, are presented in TABLE I.

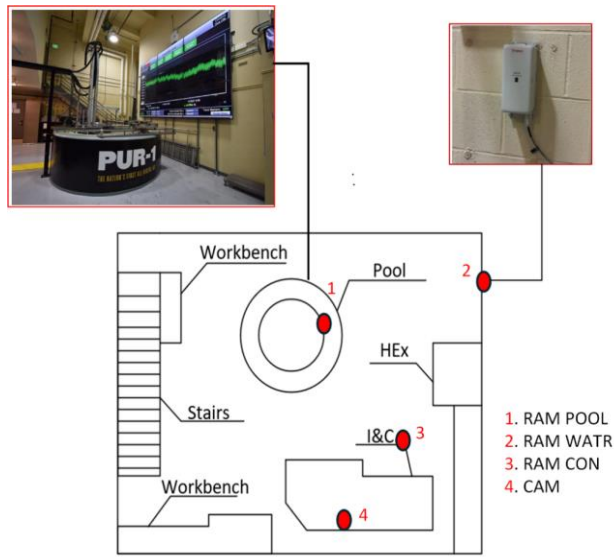

Fig. 3. Radiation detectors distribution in the reactor room.

TABLE I: Reactor Signals Used for the Training, Testing and Validation

| Feature | Units | Description |
| --- | --- | --- |
| ram pool | mR/s | Detector monitoring the top of the pool |
| ram wtr | mR/s | Detector monitoring the water infiltration system |
| ram con | mR/s | Detector monitoring the control area |
| cam | mR/s | Detector monitoring the room environment |
| neutron counts | 1/s | Neutron counts per second as measured by channel 1 |
| neutron flux | % | Neutron flux as measured by channel 4 |

10 full-power cycle datasets were collected and used for training. Each dataset contains approximately 5700 seconds (1.5 hours) of normal reactor operation, from startup to shutdown, including transients and steady state conditions at various power levels. The datasets were concatenated to create a unique dataset of approximately 50000 seconds of reactor operation. The purpose of using a large dataset is to allow the algorithm to learn the typical behavior of the radiation detectors under a variety of reactor conditions.

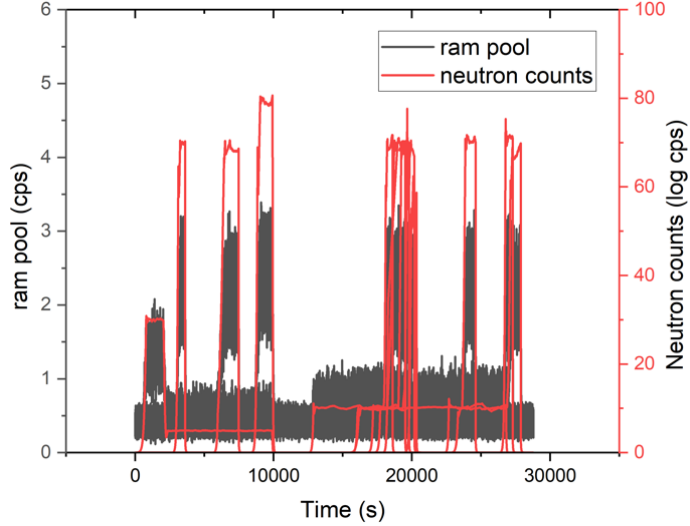

Fig. 4. Correlation between *ram pool* detector and reactor power.

### III.B Data Preprocessing

After collecting data from the monitoring software, either in real-time while the reactor is operating or at a later time by identifying operation periods, the datasets were preprocessed to ensure they were properly formatted for input into an AE model. Each dataset was normalized using a Min-Max scaler to a range between zero and one. Since the maximum values of the radiation detectors are not known, the minimum and maximum values for each feature were derived from the last three years of reactor operation to ensure consistent scaling across all datasets. To include the time dimension in the training process, a sliding window is used to capture the dynamics over a predefined window length.

The dataset, representing a variety of reactor operations, is split into training and validation sets using an 80/20 rule. Using a sliding window, which advances by one second, tensors of dimensions (batch size x window length x features) are constructed as shown in Figure 5. This format is well-suited for MTS data training, as it ensures that at each step the model considers the preceding window, reads the underlying dynamics, and learns to reconstruct the same window. The quality of the model depends on its ability to accurately reconstruct the window of data. The batch size, the window length, and other network parameters are determined during hyperparameter tuning.

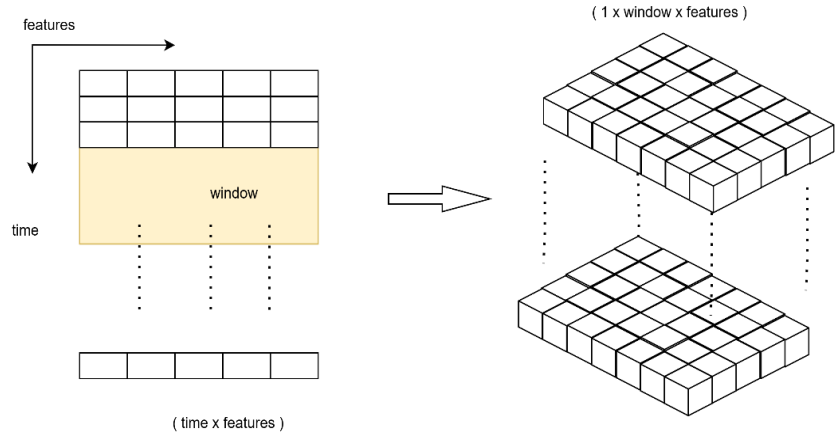

Fig. 5: Tensors created for training by sliding a window forward one second at each step.

To evaluate the performance of the trained network on new unseen data, 300 seconds of reactor recordings from a randomly selected operational mode (Figure 6a) were used to generate abnormal scenarios artificially. The abnormalities are designed to simulate irregular readings occasionally observed in the reactor, where sudden spikes occur in the sensor measurements. Three abnormal datasets of increasing complexity were created: (a) a drifted sensor, where readings in one of the sensors were altered by applying a time-varying function for the entire duration of the recording; (b) an isolated spike was inserted into an individual sensor at random position and for random duration; and (c) spikes inserted simultaneously across all four sensors, either during concurrent or overlapping time periods. The three abnormal datasets are shown in Figure 6b–d.

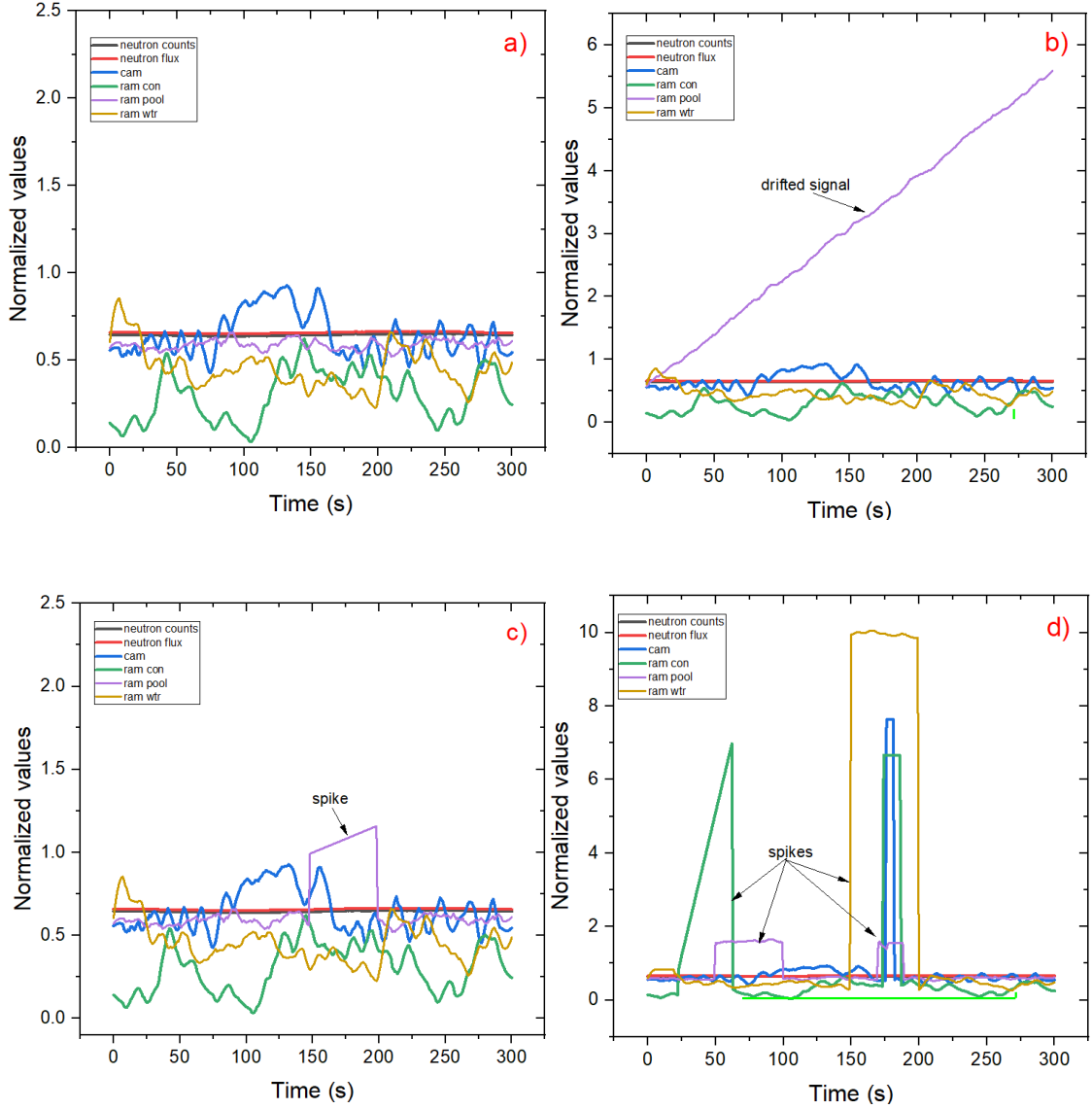

Fig. 6: Normal sensor readings (a), drifted sensor readings (b), individual spike (c), concurrent and overlapping spikes (d).

*III.C Training and Tuning*

An appropriate architecture must be defined before training any AI/ML model. The architecture is determined through hyperparameter tuning, which is a necessary step in any ML application, as the choice of hyperparameters significantly affects the model's performance [57]. Common methods for hyperparameter selection include Bayesian optimization, grid search, and random search. The tuning process considers the depth and width of the layers across the encoder, the decoder, and the latent representation. The search space spans a range of values for the hidden layers, the number of units per layer, the batch size, dropout regularization to reduce overfitting, and the learning rate.

TABLE II shows the optimal set of hyperparameters found through tuning using the random grid method. The tuner evaluated 30 different hyperparameter combinations from the predefined search space, with each model trained for 10 epochs. The criterion for the optimal selection was to minimize the Mean Squared Error (MSE) on the validation loss. The architecture that performed the best has the following structure: the input consists of sequences with shape (10x6), which are the dimensions of the tensors created in the previous step (window x features). The encoder consists of two LSTM layers, each with 64 and 32 neurons, respectively, and a dropout layer with a dropout rate of 0.06. The latent space has 64 neurons. The decoder has two hidden layers with 32 and 64 hidden units, respectively. This architecture is consistent with a typical AE structure where the encoder extracts temporal features from the sequences and reduces the dimensionality to a lower-dimensional space. The bottleneck ideally contains the most important temporal and feature-level information. The decoder reconstructs the sequences from the latent space and expands them to match the initial dimensionality. The optimal hyperparameters are presented in Table II.

After the optimal architecture is defined, the network is trained, and a model is exported for real-time testing. Figure 7 presents the training and validation losses during training for the optimal architecture.

TABLE II: Optimal hyperparameters for the AE architecture

| Learning rate | 0.006 |
|---|---|
| Batch size | 32 |
| window | 20 |
| Hidden 1 | 64 |
| Hidden 2 | 32 |
| Bottleneck | 64 |
| dropout | 0.06 |

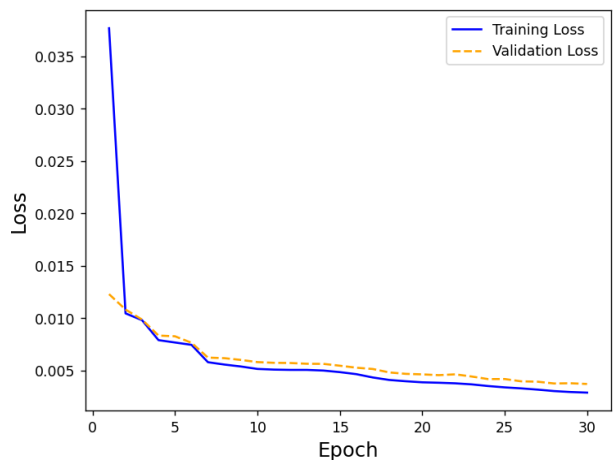

Fig. 7: Training and validation loss for the optimal architecture.

*III.D. Dual Attention Mechanism*

In this work [58] the authors propose a dual-stage attention mechanism integrated into a Recurrent Neural Network (RNN) for time series prediction. In the first stage, the feature-level attention module can adaptively identify the most relevant input features by referring to the previous hidden state. In the second stage, a temporal attention module is applied to capture dependencies across all time steps by weighting the hidden states accordingly. This dual scheme enhances both the predictive accuracy and interpretability of the model. Inspired by this work, we combine an AE with a dual-attention mechanism for the purpose of detecting and localizing potential irregularities from nuclear MTS. By training the AE with normal operational data collected across different reactor modes, a model is constructed that learns the typical behavior of a reactor system under a wide range of conditions. During inference, the AE reconstructs input sequences according to the learned representation of normal reactor behavior, and deviations from this reconstruction are flagged as potential anomalies. The dual attention mechanism further enhances the model by assigning dynamic weights, allowing it to focus on the most relevant information within each sequence.

We applied additive Bahdanau attention in both the feature and temporal spaces as extra modules during training. The feature-attention module learns to assign weights to the individual features of an input sequence, enabling the model to emphasize the most relevant features that affect the reconstruction. The temporal-attention module calculates a distribution of weights across both features and time steps, enabling the model to capture which features are most informative and at which specific steps in the sequence contribute the most or least to reconstruction. By including these modules during training, the model generates context-specific summaries of the sequence. The output is a matrix with the same dimensions as the input sequence, where higher weights are assigned to the features and time steps that have a greater influence on the reconstruction. In the context of anomaly detection from MTS data, these weights can be interpreted as highlighting the features and the duration that exhibit deviations from normal patterns for a given reactor operational mode, indicating temporal patterns or localized anomalies.

Through experimentation, it was found that applying both attention mechanisms to the latent space representation yielded the most promising results. Introducing attention at this stage offers certain advantages, as the latent space has already filtered out noise and redundant information present in the input sequence. This enables the model to focus on patterns that lie within the statistical variations of normal operation, rather than being influenced by transient fluctuations or inconsistencies in the raw data. As a result, both anomaly detection performance and model interpretability are improved. In contrast, applying the attention mechanisms in either the encoder or decoder stages produced poorer results, likely due to the presence of noisy inputs or poorly reconstructed outputs, which can distort the distribution of attention weights.

A Python package accompanies this paper and is publicly available at https://github.com/Kvasili/multianomaly-detection-attention-AE. It includes an example of running the proposed framework on the data used to simulate the abnormal events.

The training, testing, and tuning of the models were implemented using the Pytorch framework, while the data manipulation was performed with the Pandas library. All simulations were executed on a system equipped with an NVIDIA GeForce GTX 1650 Ti GPU.

## IV. RESULTS

*IV.I Normal data*

The dual attention-based AE was initially evaluated using the normal dataset (Figure 6a) to establish a baseline reference for expected sensor behavior. The MTS normal data were presented to the framework second by second, and at each time step, the algorithm considered the previous 20 seconds of data for reconstruction. The feature attention mechanism calculates weights for the six sensors, while the time-attention mechanism generates a matrix of values with the same dimensions as the input 2D sequence. After all sequences were processed, mean values were calculated for both the feature weights and the time-weights. Figure 8 shows the mean weights per feature. As observed, the weights are similar across all sensors, indicating that each sensor contributes equally to the reconstruction of the sequence. This suggests that the model correctly identifies no particular sensor as anomalous, as all readings fall within the expected statistical variability of the training data.

The temporal-attention values were also found to be relatively uniform. Figure 8 (right) presents the mean values calculated across all the 20-second sequences for the six sensors. The temporal attention values are stable and closely aligned, indicating that each time step contributes equally to the sequence. This uniform distribution suggests that there are no sudden spikes, drifts, or other abrupt changes in the data, which is consistent with normal system behavior. Applying the attention-based AE to the normal dataset establishes a baseline reference that can later be used to compare the calculated weights when analyzing the abnormal datasets.

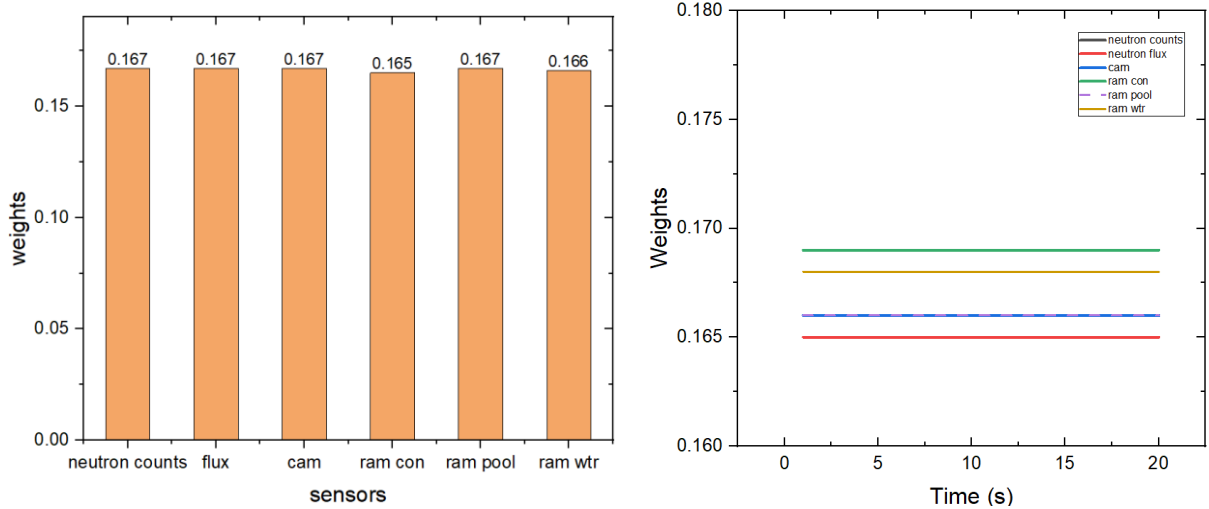

Fig. 8: Bar plot with average feature weights for the normal case (left); Average time attention per feature on the normal case (right).

*IV.II Sensor drifting*

After establishing the baseline case, the trained model was evaluated on the drifted sensor dataset (Figure 6b). In this dataset, all sensors remained within normal ranges except for the *ram pool* sensor, which was falsified to exhibit gradually drifted values over time. The dataset was presented to the framework, where at each step, the preceding 20 seconds were used to reconstruct the sequence and calculate weights for each feature and time step. A pattern emerges from analyzing the weights assigned by the attention mechanisms compared to the normal weights.

In the initial steps, when the drifted sensor values are still within the statistical variation, the weights are evenly distributed, indicating that the framework interprets the system as operating normally. As time advances and more windows are processed, the drifted values begin to deviate significantly. The feature attention consistently assigns higher weights to the falsified sensor. This indicates that the framework can detect that the specific sensor diverges from the normal behavior, signalling an anomaly in the system. For every 20-second sequence, the temporal attention matrix is also calculated to assign specific weights to each time step. The results show that all the sensors present uniformly distributed values, while the drifted sensor consistently shows lower values at all timesteps. To facilitate visualization, the mean values of both feature and temporal attention are computed over all the 20-second windows of the abnormal dataset. The results are shown in Figure 9. For the non-falsified sensors, the temporal attention values remain approximately around 0.17, whereas the drifted sensors are assigned lower values, around 0.145. This behavior indicates that the framework consistently considers the drifted sensor as less reliable at each time step, thereby reducing its influence on the sequence reconstruction process.

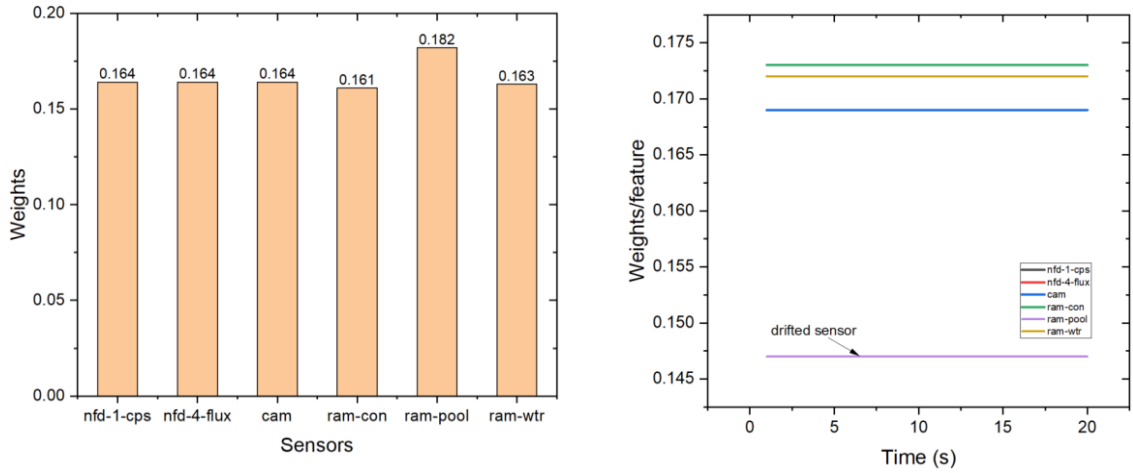

Fig. 9: Bar plot with average feature weights for the drifted sensor (left); Average time attention per feature (right).

Figure 10 presents a heatmap of the time attention for two random 20-second sequences, one taken from the beginning and the other from near the end of the time series. Initially, when the sensor drift is still minimal and close to normal values, the attention weights are more uniformly distributed across the sensors. In contrast, by the end of the time series, when the sensor drift becomes more pronounced, the heatmap clearly highlights the abnormal sensor, showing a distinct difference in its attention pattern compared to the other sensors.

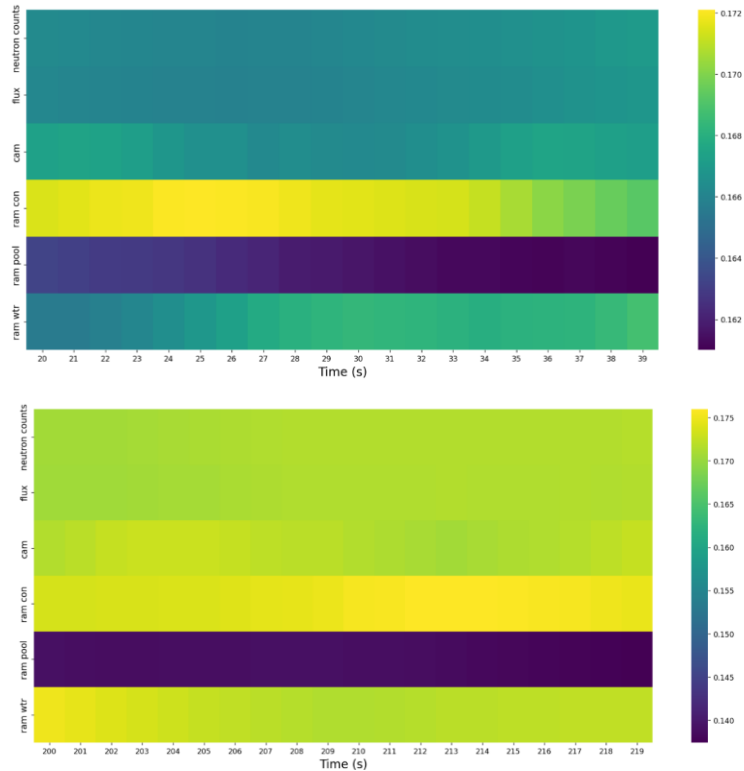

Fig. 10: Heatmap of time attention for two 20-second sequences, one taken near the beginning (left) and the other near the end (right) of the time series. The ram pool detector is clearly visualized as less important.

### *IV.III Individual Spikes*

An artificial spike was introduced between 150 and 200 seconds, with values slightly outside the normal operating range of the *ram pool* detector, as shown in Figure 6c. The nature of this spike is consistent with typical anomalies observed in the PUR-1 radiation detectors, where the values are slightly higher than normal behavior, but remain sufficiently elevated, in correlation with other detectors and signals, to indicate an anomaly. Since the algorithm processes the data with a sliding window of one second, overlapping sequences were omitted, and results are presented for the interval between 145 and 205 seconds using a 20-second sliding window. Figure 10 (left) shows the mean feature attention over the entire duration of the anomaly, while Figure 10 (right) shows the time attention results.

Consistent with the results from the other two cases, the feature attention assigns higher weights to the sensor of interest during the falsification period, while the values for the other sensors remain more evenly distributed. Since the unexpected patterns of the falsified sensor are outside the training data, the focus is more on that specific signal when reconstructing the sequence. This behavior indicates that the model identifies the sensor as potentially exhibiting an abnormal pattern. The time attention mechanism evaluates each feature to determine when it is most influential, enabling the framework to localize the anomaly within the sequence. By integrating results across consecutive sequences, the duration of the anomaly can be estimated.

The contradiction between the large values of the feature attention and the low values of the time attention can be explained by the fact that each attention mechanism addresses a different question. High feature attention values indicate that a particular feature is globally more important for the given sequence, requiring the AE to pay more attention when reconstructing the sequence. In contrast, high time attention values highlight specific time steps that are locally important. Since the falsified sensors are drifting away from normal patterns, the model assigns lower importance to their time steps, treating them as less relevant for accurate reconstruction.

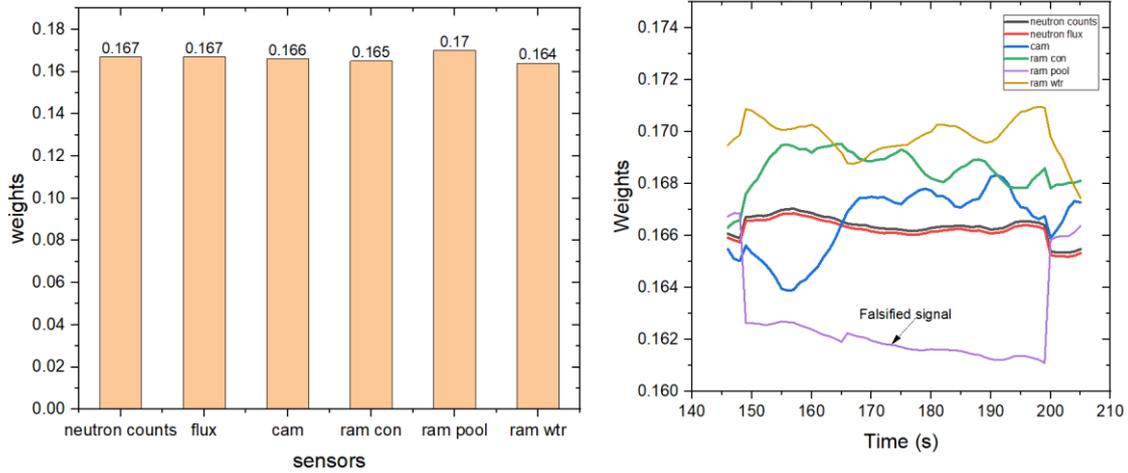

Fig. 11: Bar plot with average feature weights for the individual spike (left); Time attention per feature (right).

From the temporal weight distributions shown in Fig. 10, it is evident that the framework assigns lower importance to the ram pool sensor between 150 and 200 seconds during reconstruction. This indicates that the framework not only identifies the specific sensor exhibiting abnormal behavior but also determines the duration of the anomaly. Furthermore, Figure 12 presents two heatmaps showing the temporal weight distributions for the time sequence where the anomaly begins and the time sequence where it ends. As observed, the falsified sensor is clearly distinguishable from the other sensors. It is also observed that the model misclassifies the first three seconds and the last two seconds, corresponding to a detection accuracy of 95%.

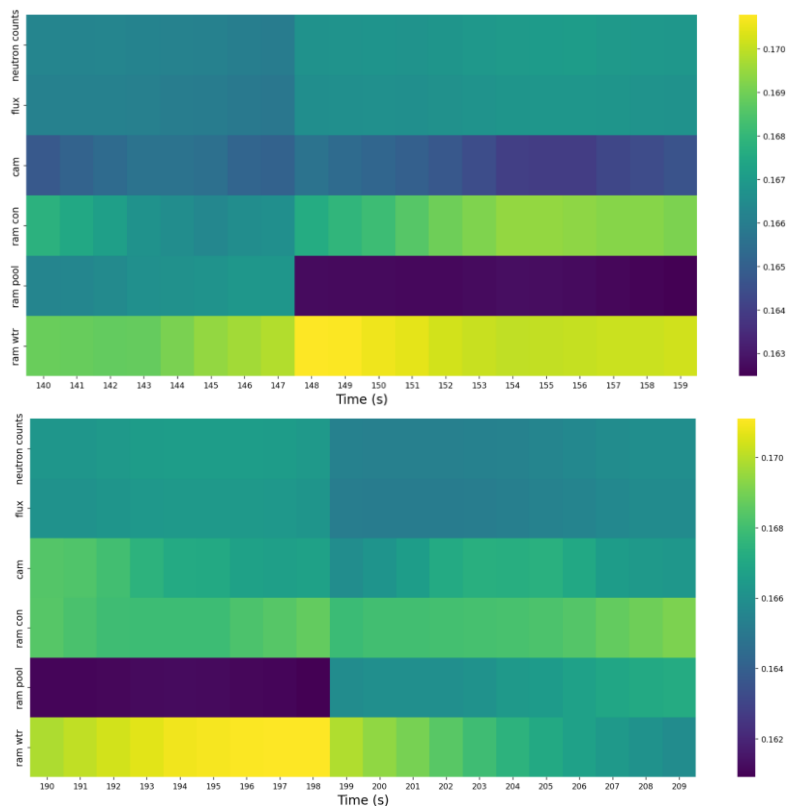

Fig. 12: Heatmaps showing the temporal weight distributions for the time intervals where the anomaly begins and ends.

## IV.IV Concurrent – overlapping spikes

The final case presents greater challenges compared to the other three cases. For this scenario, the testing dataset was infused with abnormal values in a pattern that allowed individual, overlapping, and concurrent anomalies to occur. This setup aims to simulate conditions in which multiple sensors exhibit anomalies simultaneously or in overlapping intervals. The distribution of anomalies throughout the dataset is shown in Figure 6d.

To avoid complications caused by overlapping windows, the results are reported using a sliding window of 20 seconds. For clarity, we present results for two cases: (1) a single 20-second window containing multiple anomalies, and (2) the entire duration of the testing dataset, which includes all abnormalities. Figure 13 shows the results for the interval between 180 and 200 seconds. In this timeframe, the *ram wtr* sensor is falsified for the entire 20-second period, *ram con* is falsified for approximately half of the period, and the *cam* sensor is falsified only during the first four seconds. The feature attention assigns the highest weight value to the sensor with the most extensive falsification (1.91 for *ram wtr*), moderately high weight to the next sensor (1.17 for *ram con*), and slightly higher values (1.62) for the sensor affected for the shortest duration (*cam*). The non-falsified sensors present the lowest weight values. The time attention displays lower values precisely at the timesteps where falsifications occur. These results indicate that the model behaves as intended, aligning with previous cases.

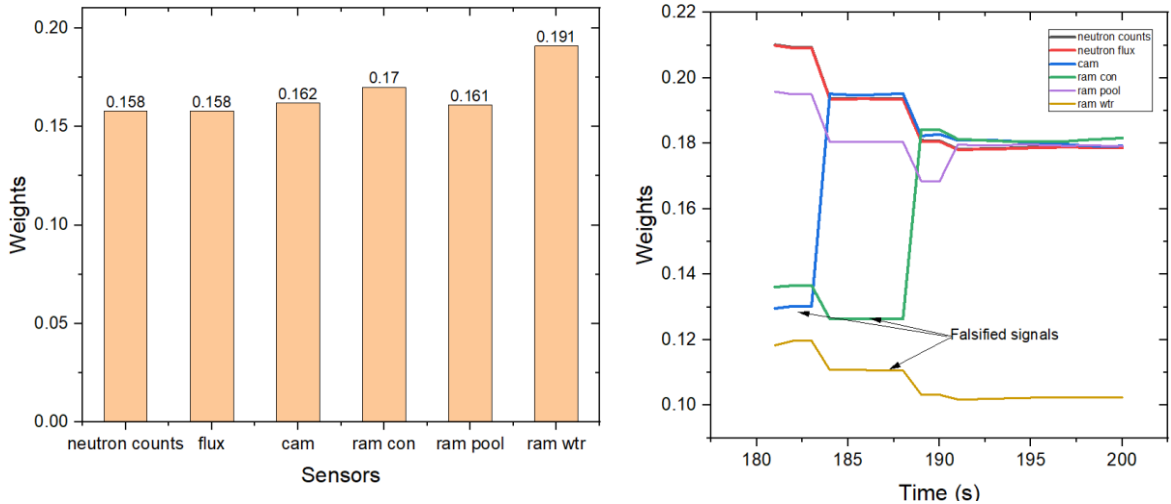

Fig. 13: Feature weights (left); Time attention per feature (right).

To gain a better understanding of the AE's performance across the entire testing dataset, the temporal attention results for the full duration are presented in Figure 14. For comparison, the actual time series is also presented. The non-falsified sensors exhibit concentrated values, indicating that the model does not assign specific attention to them. In contrast, the falsified sensors are assigned lower weights. A clear inverse relationship is observed where the higher the value of the falsified sensor, the lower the weight assigned by the model. This behavior shows that the model can successfully distinguish between individual, overlapping, and concurrent falsifications and provide evidence of the model's ability to achieve clear temporal localization of anomalies.

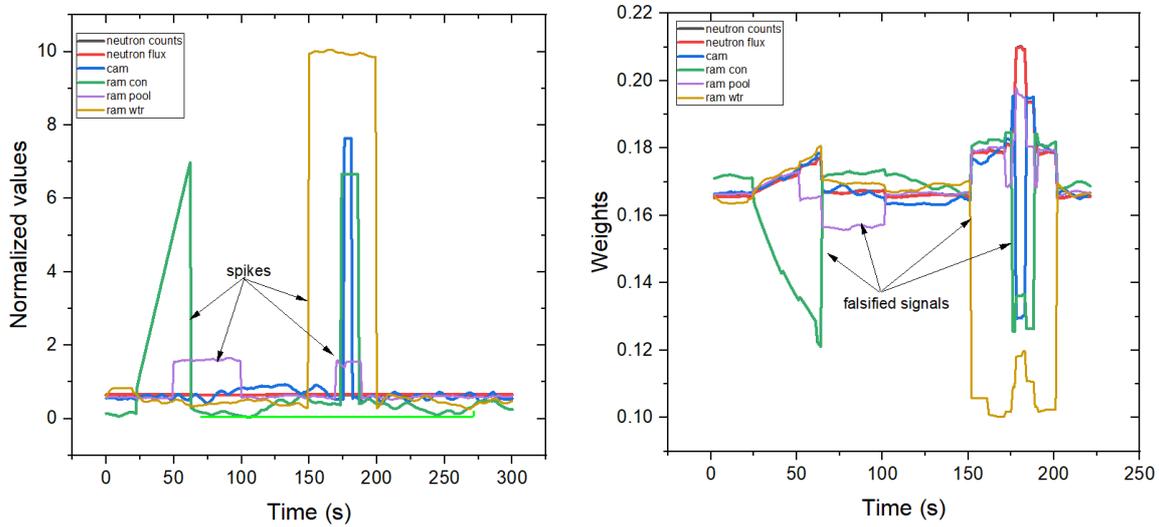

Fig. 14: Concurrent and overlapping anomalies (left). Weight distribution for temporal attention per feature (right).

Figure 15 shows a heatmap of the temporal weight distribution for the time interval between 160 and 180 seconds. The visualization effectively highlights the periods during which the sensors were falsified in the sequence. Interestingly, while the *ram pool* detector was infused with slightly higher than normal values, the neutron counts simultaneously exhibited elevated values. The system interprets this correlation as being within normal boundaries and, therefore, assigns normal temporal weights to the *ram-pool* detector for the period of correlation (around 175 seconds). This condition does not apply earlier in the time series (around 50 seconds), where the slightly elevated values of the *ram pool* detector are identified as abnormal due to the absence of a corresponding increase in reactor power.

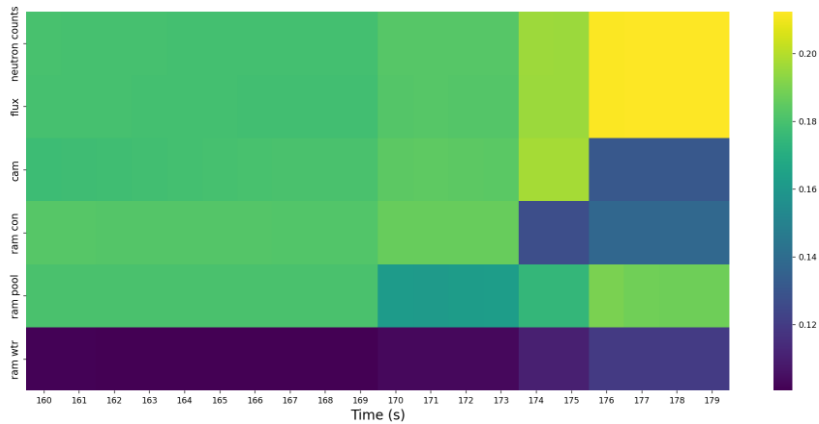

Fig. 15: Heatmap with temporal attention for the time interval between 160-180 seconds

Lastly, the results are consistent with the previous scenarios. In all cases, the feature attention identified which sensors were compromised, while the time attention pinpointed when anomalies occurred and their duration. When a sudden spike or falsification is present, the feature attention assigns higher weights to the affected sensors. This behavior likely arises because the AE recognizes these sensors as the primary sources of deviation in the system and therefore focuses on them during reconstruction. Even with overlapping and concurrent falsifications, the attention mechanism allocates higher global weights to the compromised sensors. In contrast, the time attention reflected the model's confidence in its reconstruction by assigning lower values to falsified sensors and higher values to non-falsified ones. These two attention mechanisms complement each other. By examining them together, it is possible to distinguish between clear irregularities and more subtle deviations.

**V. CONCLUSIONS**

In this work, we develop and investigate an unsupervised end-to-end LSTM-based AE architecture with a dual attention mechanism applied to the latent representation for detecting and localizing radiation abnormalities in sensors from the PUR-1 research reactor. The network is trained exclusively on normal reactor data collected under various operating conditions to establish a baseline model of normal reactor behavior, without requiring prior knowledge of the system or potential abnormalities. Synthetic abnormalities were introduced to simulate irregularities observed in the reactor system, with increasing levels of complexity, ranging from gradual sensor drifts to overlapping and concurrent faults. The weights assigned by the dual attention mechanism can be interpreted as sensitivity coefficients, highlighting subsets of the dataset that have the greatest impact on the system's internal dynamics, and therefore serving as indicators of feature importance in the system process. The results across all test scenarios show that the dual-attention AE can both distinguish between affected sensors and accurately localize anomalies in time.

Although the framework was benchmarked on real-world datasets from PUR-1, it is estimated that it can be applied to a wide range of practical scenarios where the normal behavior of a control system is known in the form of MTS data, and can be used for training. As future work, we aim to investigate how the number of falsified sensors affects the performance of the attention-based AE. Additionally, the framework is planned to be integrated into the current control system of PUR-1, where its autonomous operation will be evaluated. Lastly, recent studies [59] [60] have explored the use of SHAP analysis to provide explainability in anomaly detection as part of cybersecurity modules for networked reactor systems. Building on this, we plan to examine whether the proposed attention mechanism can also be leveraged to detect and localize cybersecurity threats within the PUR-1 infrastructure.

ACKNOWLEDGMENTS

This project was funded by DOE Office of Nuclear Energy's Nuclear Energy University Program under contract DE-NE0009268.

DISCLOSURE STATEMENT

All of the authors affirm that there are no conflicts of interest to report for this study.